\documentclass[letterpaper]{article} 
\usepackage{aaai2027}  
\usepackage[hyphens]{url}  
\usepackage{graphicx} 
\usepackage{natbib}  
\usepackage{caption} 
\usepackage{algorithm}
\usepackage{algorithmic}
\usepackage{booktabs}
\usepackage{multirow}
\usepackage{amsmath}
\usepackage{amsfonts}
\usepackage[table,xcdraw]{xcolor}
\definecolor{no1}{HTML}{FFB2B2}  
\definecolor{no2}{HTML}{FFD9B2}  

\usepackage{newfloat}
\usepackage{listings}
\DeclareCaptionStyle{ruled}{labelfont=normalfont,labelsep=colon,strut=off} 
\floatstyle{ruled}
\newfloat{listing}{tb}{lst}{}
\floatname{listing}{Listing}

\newcommand{\projecttitle}{DUET4D}

\usepackage{booktabs}

\title{\projecttitle: Self-Corrective Coupled Guidance for Sparse-View 4D Reconstruction}
\author{
    Beizhen Zhao\textsuperscript{\rm 1}\equalcontrib,
    Nanjie Yao\textsuperscript{\rm 1}\equalcontrib,
    Boyi Fu\textsuperscript{\rm 1},
    Qingsong Yan\textsuperscript{\rm 2}, 
    Haofan Ren\textsuperscript{\rm 2},
    Kaiyong Zhao\textsuperscript{\rm 2}, 
    Hao Wang\textsuperscript{\rm 1}\corresponding
}

\affiliations{
    \textsuperscript{\rm 1}The Hong Kong University of Science and Technology (Guangzhou)\\
    \textsuperscript{\rm 2}XGRIDS\\

    bzhao610@connect.hkust-gz.edu.cn,
    nyao656@connect.hkust-gz.edu.cn,
    haowang@hkust-gz.edu.cn
}

\makeatletter
\let\@oldmaketitle\@maketitle
\renewcommand{\@maketitle}{\@oldmaketitle
  \begin{center}
  \vspace{-4em} 
    \captionsetup{type=figure}
    \includegraphics[width=1.0\textwidth, trim=0cm 12.1cm 9.2cm 0cm, clip]{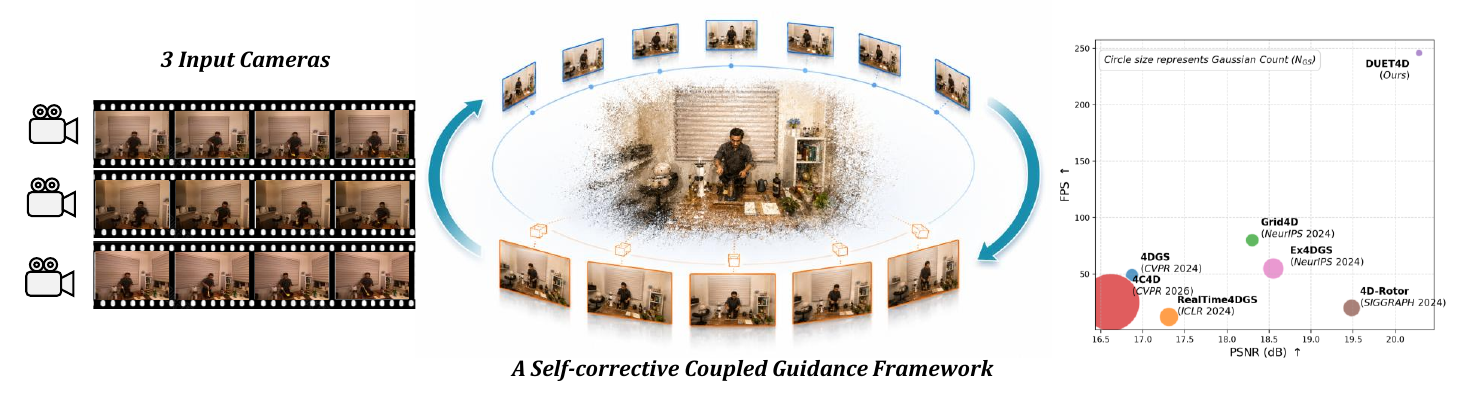}
    \captionof{figure}{We present \textbf{DUET4D}, a self-corrective coupled guidance framework for Sparse-View 4D Reconstruction. Compared with existing methods, we achieve better rendering quality with substantially higher rendering FPS and a more compact Gaussian representation.}
    \label{fig:teaser}
  \end{center}
}
\makeatother

\begin{document}

\maketitle

\begin{abstract}
Sparse-view 4D reconstruction is highly under-constrained when only a few cameras are available. 
Generative priors can recover missing appearances, but video models are computationally impractical for iterative reconstruction, whereas single-view models produce temporally inconsistent pseudo-observations. 
Moreover, directly absorbing uncertain generative outputs into Gaussian optimization often creates redundant primitives and floating artifacts.
To address these challenges, we propose \projecttitle, a self-corrective coupled guidance framework for sparse-view 4D Gaussian splatting. 
The core of our approach is a self-corrective framework where the 4D representation guides generative prior for spatio-temporal consistent supervisions. 
We first construct sequential alignments from current Gaussian evidence to steer the latent restoration. 
Then we use Sequential Fidelity Resampling to lift generative supervisions into Gaussian space to selectively distinguish repeatedly supported primitives.
The corrected representation then provides more reliable Gaussian evidence for the next restoration cycle.
Extensive experiments demonstrate that \projecttitle~significantly improves rendering quality and rendering efficiency.
\end{abstract}

\section{Introduction}
By extending 3D Gaussian Splatting to the temporal domain, 4D Gaussian methods can reconstruct and render dynamic scenes with high visual quality~\cite{cai2022neural,du2021neural,fang2022fast,kratimenos2024dynmf,li2022neural}.
However, most existing methods rely on sufficiently dense multi-view observations~\cite{shi2026sparse4dgs,pan2026sparsecam4d, zhou20264c4d}.
When the input is reduced to only a few cameras, reconstruction quality degrades severely.
Many regions of the scene are never directly observed, while dynamic objects are only partially constrained across time.
As a result, sparse-view 4D Gaussian reconstruction often suffers from distorted artifacts, floating Gaussians, and temporally inconsistent motion.
Existing works~\cite{shi2026sparse4dgs, zhou20264c4d} introduce texture regularization and Gaussian decaying to constrain sparse-view reconstruction. 
However, they are insufficient to recover stable dynamic appearance under large viewpoint drifts.

To introduce stronger generative priors, SparseCam4D~\cite{pan2026sparsecam4d} utilizes video diffusion models (VDM) to synthesize pseudo-observations, but iteratively querying a VDM is computationally impractical for multi-view 4D optimization. 
This presents the first challenge: single-view diffusion models (SDMs) provide a more efficient alternative, yet their independent predictions lack cross-view and temporal awareness, leading to temporal incoherent motions.
However, temporal coherence alone does not make pseudo-supervision uniformly reliable.
There remains a spatial drift between the generative supervisions and the sparse observations.
Blindly densifying these locations leads to redundant primitives and floating Gaussians.
A second challenge is therefore to selectively lift pseudo supervision into reliable representation-side decisions.

To address these challenges, we propose \projecttitle, a self-corrective framework that couples a frozen single-view diffusion model with an evolving 4D Gaussian representation. 
Instead of relying on a computationally heavy VDM or directly using independently restored views as supervision, DUET4D exploits the current 4D Gaussians as a spatiotemporal correspondence carrier. 
The resulting Gaussian-derived alignments and confidence estimates construct a reliable anchor that steers the latent denoising direction of a frozen SDM, enabling temporally coherent pseudo-observations without training or invoking an additional video model.

Crucially, DUET4D does not indiscriminately absorb these generated observations into the representation. 
The Sequential Fidelity Resampling distinguishes reliable reconstruction deficits from persistent inconsistencies by jointly considering confidence-gated pseudo supervisions and real observations, and then lifts these 2D signals to individual 3D primitives. 
The updated representation consequently provides cleaner correspondence evidence for the next restoration round, forming a closed self-corrective cycle that improves reconstruction fidelity while suppressing redundant Gaussians.

Our contributions are summarized as follows:
\begin{itemize}

\item We propose \projecttitle, a unified framework for sparse-view 4D reconstruction, establishing a self-corrective framework where the 4D representation guides generative prior for spatio-temporal consistent supervisions.

\item We introduce a training-free anchor-driven latent steering mechanism that uses Gaussian-derived correspondences to make a frozen SDM produce temporally coherent pseudo-supervisions.

\item We design Sequential Fidelity Resampling to selectively lift confidence-supported pseudo-view discrepancies into Gaussian densification while pruning primitives that remain inconsistent with real observations, producing a cleaner and more compact representation.
    
\end{itemize}

\section{Related Work}

\paragraph{Sparse-view Dynamic Reconstruction}

Dynamic scene reconstruction requires temporally coherent modeling of appearance and motion. 
While NeRF-based methods rely on time-conditioned fields or feature factorization~\cite{mildenhall2021nerf,pumarola2021d,park2021nerfies,park2021hypernerf,cao2023hexplane,fridovich2023k},4DGS-based approaches significantly improve efficiency via deformation fields, tracking, or direct 4D representations~\cite{kerbl20233d,yang2024deformable,
li2024spacetime,luiten2024dynamic,lin2024gaussian,wu20244d,yang2024real,duan20244d}. 
However, these methods typically require dense, synchronized multi-view videos and degrade heavily under sparse camera setups.
To tackle the under-constrained sparse-view setting, prior works exploit category-specific priors like SMPL for humans~\cite{peng2021animatable,peng2021neural,weng2022humannerf,hu2024gauhuman} or monocular depth and tracking cues~\cite{lei2025mosca,wang2025shape,pradeep2013monofusion}. 
Though providing useful temporal constraints, recent methods~\cite{shi2026sparse4dgs,zhou20264c4d}still suffer from unstable appearance under large viewpoint changes.

\paragraph{Diffusion Models for Novel-View Synthesis} Diffusion models serve as powerful priors for 3D/4D generation. While object-centric models~\cite{liu2023zero,liu2024syncdreamer,shi2024mvdream,long2024wonder3d} achieve strong multi-view consistency, they cannot directly handle scene-level dynamic reconstruction. 
Recent camera-controlled video diffusion models~\cite{wang2024motionctrl,he2024cameractrl,xu2024camco,yu2024viewcrafter,bai2025recammaster,wu2025cat4d} synthesize novel-view dynamic observations, offering richer supervision. 
However, they rely on expensive multi-frame denoising and often produce cross-view or temporal inconsistencies that harm 4D optimization.
To incorporate diffusion into sparse dynamic reconstruction, recent methods such as Diffuman4D~\cite{jin2025diffuman4d} and SparseCam4D~\cite{pan2026sparsecam4d} query video diffusion models for better rendering quality, which take ~40 hours for a single dynamic scene, facing a strict trade-off between consistency and efficiency. 
We address this through a unified framework that constructing a self-corrective coupled pipeline where the 4D representation guide generative prior for spatio-temporal consistent supervisions.

\begin{figure*}[t]
    \centering
    \includegraphics[width=1.0\textwidth, trim=0cm 7.6cm 12.5cm 0cm, clip]{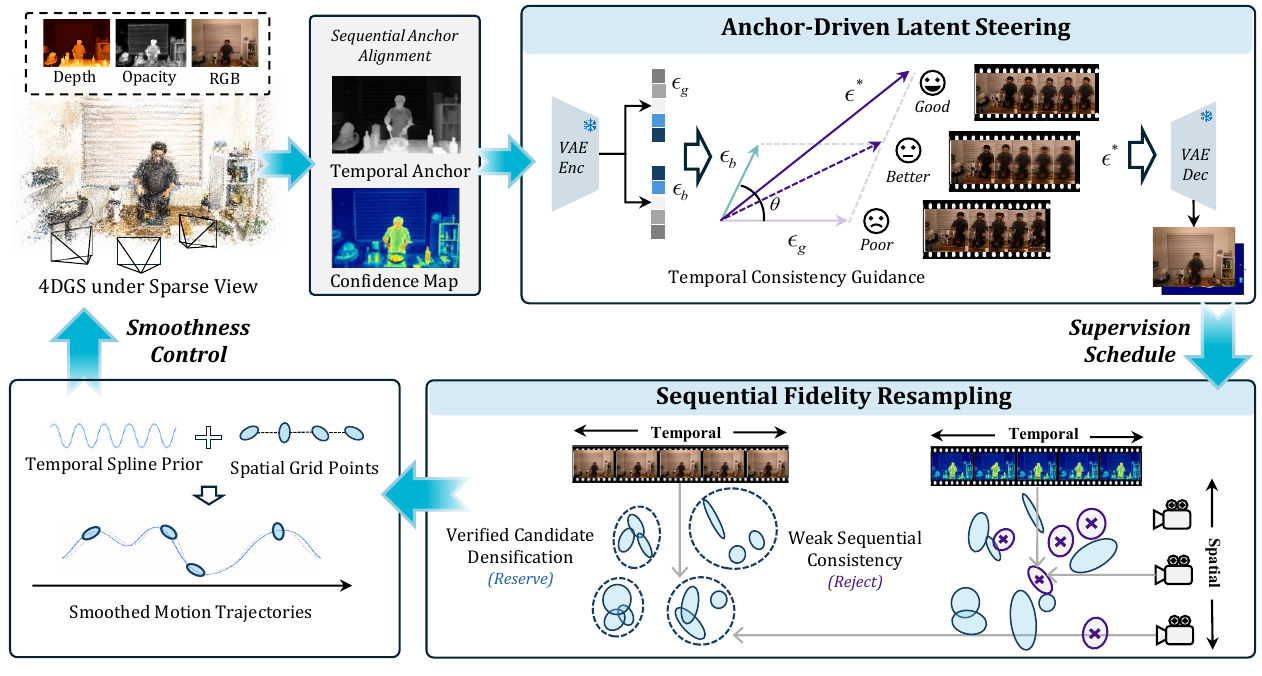}
    \vspace{-0.3cm}
    \caption{\textbf{Overview of \projecttitle.} First, we construct sequential anchor alignment through Gaussian evidence to generate temporal anchor frames and confidence map.
    Then we design a training-free anchor-driven latent steering mechanism within the latent space of SDM for temporally consistent generative outputs.
    The frames and confidence map will be used to locate Gaussians with weak spatial and temporal consistency for optimization.
    The reserved Gaussians are controlled by a spatial grid and Fourier spline priors for motion smoothness and regularization.}
    \label{fig: overview}
    \vspace{-0.3cm}
\end{figure*}

\section{Methodology}
\label{method}

\subsection{Sequential Anchor Alignment}

Independent single-view restoration can recover missing appearance details, but its correction is estimated separately for each virtual view and therefore does not constitute reliable 4D supervision. Rather than introducing an additional video model or correspondence network, we use the evolving 4D Gaussian representation itself to determine which restoration changes are supported across nearby views and time. At each restoration round, a virtual node $i=(v,t)$ provides the current Gaussian rendering $x_i$ and its independently restored output $o_i$. We retain only the bounded restoration residual $r_i=\operatorname{clip}(o_i-x_i,-\kappa_r,\kappa_r)$, so that the alignment transfers appearance corrections instead of directly replacing the geometry, motion, and view-dependent content already encoded by the current representation.

For a neighboring node $j\in\mathcal{N}(i)$, we track the same canonical Gaussians after deformation to construct a target-indexed backward sampling grid $W_{j\rightarrow i}$. The reliability of this representation-derived correspondence is measured by
\begin{equation}
\begin{split}
c_{j\rightarrow i}(p) ={}& \mathbb{I}_{\mathrm{valid}}(p)
s\!\left(\alpha_i(p)\right)
s\!\left(W_{j\rightarrow i}(\alpha_j)(p)\right) \\
&\cdot \exp\!\left(
-\frac{e_{\mathrm{dep}}(p)}{\tau_{\mathrm{dep}}}
-\frac{e_{\mathrm{cyc}}(p)}{\tau_{\mathrm{cyc}}}
-\frac{e_{\mathrm{pho}}(p)}{\tau_{\mathrm{pho}}}
\right).
\end{split}
\label{eq:soft_confidence}
\end{equation}
where $\mathbb{I}_{\mathrm{valid}}$ rejects invalid projections and $s(\alpha)$ suppresses low-opacity correspondences. The depth, cycle, and photometric terms jointly exclude mappings that are geometrically unsupported or inconsistent with the current representation. Consequently, all alignment evidence is obtained from the evolving 4DGS without requiring external optical flow, monocular depth, or video diffusion.

We then align neighboring restoration residuals rather than their complete RGB predictions. A neighbor contributes only when its warped residual agrees with the target restoration:
\begin{equation}
\begin{aligned}
a_{ji}(p)
={}&\lambda_{ji}c_{j\rightarrow i}(p)
\exp\!\left(
-\frac{
\left|W_{j\rightarrow i}(r_j)(p)-r_i(p)\right|_1
}{\sigma_r}
\right),\\
\bar r_i(p)
={}&
\frac{
\sum_{j\in\mathcal{N}(i)}
a_{ji}(p)W_{j\rightarrow i}(r_j)(p)
}{
\sum_{j\in\mathcal{N}(i)}a_{ji}(p)+\delta
}.
\end{aligned}
\end{equation}

To handle pixels with insufficient support near occlusions, motion boundaries, or image borders, we jointly define the accumulated support and its bounded confidence mask as
\begin{equation}
\begin{aligned}
A_i(p)
&=
\frac{1}{\lvert\mathcal{N}(i)\rvert}
\sum_{j\in\mathcal{N}(i)}a_{ji}(p),\\
m_i(p)
&=
\operatorname{clip}\!\left(
\frac{A_i(p)}{1+A_i(p)},0,m_{\max}
\right).
\end{aligned}
\label{eq:trust_mask}
\end{equation}
Here, $m_{\max}\leq1$ bounds the trust assigned to neighboring residuals. The mask is zero without valid support and increases with accumulated correspondence confidence.

Using $\kappa_c$ to bound the correction proposed by neighboring residuals, we directly construct the spatiotemporal anchor as
\begin{equation}
x_i^{\mathrm{st}}(p)
=
x_i(p)+r_i(p)
+\operatorname{clip}\!\left(
\bar r_i(p)-r_i(p),-\kappa_c,\kappa_c
\right).
\end{equation}
All independent outputs $o_i$ are computed before aggregation, and no aggregated residual is reused. The construction is therefore independent of graph traversal order and avoids recursive error propagation within one aggregation round.

\subsection{Anchor-Driven Latent Steering}
\label{sec:dual_path_diffusion}

Given a rendered image and its fixed restoration configuration, the restorer performs a single denoising step in the latent space and produces two outputs: an internal model prediction $\epsilon$ and a decoded restored RGB image $o$. 
The internal prediction is used for dual-path latent steering, whereas the decoded RGB image is used only to construct the final pseudo-observation.

Specifically, we run the same frozen SDM restorer through two parallel paths: a base path and an anchor-guided path. 
The base and anchor-guided paths take $x_i$ and $x_i^{\mathrm{st}}$ as input, respectively, yielding internal predictions $\epsilon_b$ and $\epsilon_g$. 
Both predictions share the condition and the parameters.
To extract a confidence-aware guidance signal, we compare the two prediction directions under the mask $m_i$ constructed in Eq.~\ref{eq:trust_mask}.
This mask reflects the local 4DGS-supported anchor correspondence consistency, so the anchor-guided path is trusted mainly in regions with strong valid support. 
Inspired by previous works~\cite{ho2021classifier,song2025worldforge}, we obtain the guided prediction as
\begin{equation}
\epsilon^{*}=\epsilon_g+\rho\cdot\sin\theta\left(
\epsilon_g-
\frac{\lVert \epsilon_g\rVert_{m}}
{\lVert \epsilon_b\rVert_{m}+\delta}
\cos\theta\cdot\epsilon_b
\right),
\label{eq:orthogonal_guidance}
\end{equation}
where $\cos\theta =\operatorname{clip}\left(\langle\epsilon_g,\epsilon_b\rangle_m /       (\lVert\epsilon_g\rVert_m\cdot\lVert\epsilon_b\rVert_m+\delta), -1, 1 \right)$ is the cosine-similarity  between the two predictions, while $\sin\theta = \sqrt{1-\cos^2\theta}$ is the sine counterpart. $\lVert\cdot\rVert_m$ denotes the weighted norm, and $\rho$ controls the guidance strength. 

The parenthesized term extracts the confidence-weighted component of $\epsilon_g$ orthogonal to $\epsilon_b$ after magnitude alignment, while $\sin\theta$ scales the guidance by the orthogonal discrepancy. 
The guidance encourages spatiotemporal consistency in well-supported regions while preserving local restoration details and limiting the influence of unsupported correspondences.

The guided prediction $\epsilon^{*}$ is decoded into a guided restoration $\tilde{o}_i$. Although the confidence mask has already been used to estimate a guidance direction, the resulting restoration may still contain uncertain content in regions affected by occlusions, motion boundaries, or inaccurate correspondences. We therefore apply the confidence mask once more in RGB space to selectively integrate the guided restoration:
\begin{equation}
y_i=o_i+m_i\odot\left(\tilde{o}_i-o_i\right).
\label{eq:rgb_confidence_fusion}
\end{equation}
Supported regions thus adopt the spatiotemporally guided restoration, whereas uncertain regions revert to the independently restored result $o_i$,
preventing unsupported neighboring evidence from being explicitly fused into the final pseudo-observation. Finally, the fused output and its confidence mask serve as pseudo-supervision and confidence input, respectively, for optimizing the 4DGS representation.

\subsection{Sequential Fidelity Resampling}

Although the generative supervisions $y_i$ are temporally more coherent, they are still image-space predictions with spatially varying reliability. 
Directly treating every large 2D gradints as a densification signal would consequently absorb generative uncertainty into the explicit representation. 
To mitigate this ambiguity, we introduce Sequential Fidelity Resampling, which evaluates the current Gaussians in two complementary evidence domains.

At the $r$-th resampling cycle, we sample a set of virtual-camera nodes $\mathcal{S}_{\mathrm{ps}}$ and render the current Gaussian representation as $x_i$ at each node $i=(v,t)$. Given the fused pseudo-observation $y_i$ and the correspondence confidence $m_i$ inherited from Secs.~3.1 and~3.2, we first identify reconstruction deficits as
\begin{equation}
b_i^{\mathrm{ps}}(p)
=
\mathbb{I}\!\left[
e_i^{\mathrm{ps}}(p)>\tau_e
\;\land\;
m_i(p)>\tau_m
\right],
\label{eq:pseudo_deficit}
\end{equation}
where $e_i^{\mathrm{ps}}(p)=\left\|x_i(p)-y_i(p)\right\|_1$ denotes the photometric discrepancy. 
The confidence gate suppresses residuals around unsupported motion boundaries, so that pseudo-view discrepancies are used only when they are supported by the current spatiotemporal correspondence.

Let $a_{ik}(p)$ denote the rasterized contribution of Gaussian $g_k$ to pixel $p$ at node $i$. We lift the confidence-gated residuals into Gaussian space and define the pseudo-view support as
\begin{equation}
S_{k,r}^{\mathrm{ps}}=\frac{1}{|\mathcal{S}_{\mathrm{ps}}|}\sum_{i\in\mathcal{S}_{\mathrm{ps}}}\frac{\sum_{p}a_{ik}(p)b_i^{\mathrm{ps}}(p)e_i^{\mathrm{ps}}(p)}{\sum_{p}a_{ik}(p)+\delta}.
\label{eq:pseudo_support}
\end{equation}
Since the score is accumulated over multiple spatial and temporal nodes, a high $S_{k,r}^{\mathrm{ps}}$ indicates that $g_k$ repeatedly contributes to reliable under-reconstructed regions.

We separately evaluate the same representation using the real observations. Let $\mathcal{S}_{\mathrm{obs}}$ denote the sparse input-camera nodes and $I_i$ denote the corresponding ground-truth image. Their photometric discrepancy is
$e_i^{\mathrm{obs}}(p).$
For each Gaussian that is sufficiently visible in the observed views, we compute
\begin{equation}
E_{k,r}^{\mathrm{obs}}=\frac{\sum_{i\in\mathcal{S}_{\mathrm{obs}}}\sum_{p}a_{ik}(p)e_i^{\mathrm{obs}}(p)}{\sum_{i\in\mathcal{S}_{\mathrm{obs}}}\sum_{p}a_{ik}(p)+\delta},
\end{equation}
A high $E_{k,r}^{\mathrm{obs}}$ provides direct evidence that a visible primitive remains incompatible with the sparse real observations.

We use the two accumulated signals to regulate Gaussian topology. 
A primitive is selected for pruning only when it is not supported by persistent fidelity evidence. 
Specifically, we define
\begin{equation}
\mathcal{P}_r=\left\{g_k\,\middle| \left(E_{k,r}^{\mathrm{obs}}>\tau_p\right)\land\left(S_{k,r}^{\mathrm{ps}}<\tau_p\right)\right\}.
\label{eq:pruning_set}
\end{equation}
The prune policy considers both real observations and generative supervisions.
This prevents such residuals from being repeatedly interpreted as missing geometry.

For the remaining primitives, we therefore define the densification set as
\begin{equation}
\mathcal{D}_r=\left\{g_k\,\middle|\,g_k\notin\mathcal{P}_r,\ S_{k,r}^{\mathrm{ps}}>\tau_d,\ \right\}.
\label{eq:densification_set}
\end{equation}
Sequential Fidelity Resampling therefore uses the pseudo-view space to propose repeatedly supported reconstruction deficits and the observed-view space to reject contradictory primitives, selectively lifting generative supervision into Gaussian-space densification and pruning rather than blindly following image-space residuals.

\subsection{Smoothness Control Regularization}
To avoid locally inconsistent motion under sparse observations, we therefore apply a smoothness control after resampling stage to regularize the motion of the active Gaussians.
We partition the canonical space using a multi-resolution hash grid feature $\mathbf{h}_{\mathrm{sp}}(\boldsymbol{\mu}_k)$ for each Gaussian center $\boldsymbol{\mu}_k$. 
\begin{equation}
\mathcal{L}_{\mathrm{sp}}=\mathbb{E}_{k,\Delta\boldsymbol{\mu}}\left[\left\|\mathbf{h}_{\mathrm{sp}}(\boldsymbol{\mu}_k+\Delta\boldsymbol{\mu})-\mathbf{h}_{\mathrm{sp}}(\boldsymbol{\mu}_k)\right\|_2^2\right].
\label{eq:spatial_smoothness}
\end{equation}
This constraint encourages neighboring locations to produce compatible deformation features while preserving motion discontinuities between sufficiently separated grid regions.

For each active Gaussian, we use a set of spline basis functions $\mathbf{h}_{\mathrm{tmp},k}(t)$ to provide a continuous temporal trajectory and Smoothness Regularization:
\begin{equation}
\mathcal{L}_{\mathrm{tmp}}=\mathbb{E}_{k,t}\left[\left\|\mathbf{d}_k(t+\Delta t)-2\mathbf{d}_k(t)+\mathbf{d}_k(t-\Delta t)\right\|_2^2\right],
\label{eq:temporal_smoothness}
\end{equation}
where $\mathbf{d}_k$ denotes temporal deformations with spline prior.
This term preserves non-rigid displacement while reducing local trajectory jitter.

The corrected and motion-regularized representation is subsequently used to render the next spatiotemporal anchor graph, providing cleaner correspondence evidence for the following restoration cycle and thereby closing the self-corrective coupled of DUET4D.

\newcommand\rowrule{\rule{0pt}{10pt}}

\begin{table*}[t]   
    \caption{\textbf{Quantitative comparisons on the Nvidia Dynamic Scenes~\citep{yoon2020novel}, Neural 3D Video~\citep{li2022neural}, and Google Immersive~\citep{broxton2020immersive} datasets.} 
    The \textbf{best} and \underline{second-best} results are highlighted in bold and underlined, respectively. Our method outperforms all baseline methods across all evaluated metrics. ``$\uparrow$'' indicates higher is better, while ``$\downarrow$'' indicates lower is better. Specifically, DUET4D$^{*}$ uses exactly the same preprocessed inputs as SparseCam4D, while all model components and optimization settings remain unchanged. }
    \vspace{-1mm}
    \centering
    \resizebox{1.0\linewidth}{!}{%
    \begin{tabular}[t]{l | c |ccccc|ccccc|ccccc}
        \toprule[2pt]
        \multirow{2}{*}{Method} & \multirow{2}{*}{Venue}
        & \multicolumn{5}{c|}{Nvidia Dynamic Scenes}
        & \multicolumn{5}{c|}{Neural 3D Video}
        & \multicolumn{5}{c}{Google Immersive}\\
        & 
        & FPS$\uparrow$ & $N_{\text{GS}}\downarrow$
        & PSNR$\uparrow$ & SSIM$\uparrow$ & LPIPS$\downarrow$
        & FPS$\uparrow$ & $N_{\text{GS}}\downarrow$
        & PSNR$\uparrow$ & SSIM$\uparrow$ & LPIPS$\downarrow$
        & FPS$\uparrow$ & $N_{\text{GS}}\downarrow$
        & PSNR$\uparrow$ & SSIM$\uparrow$ & LPIPS$\downarrow$\\
        \midrule

        \rowrule 4DGS & CVPR 2024
        & 54&\underline{0.26}M&16.28&0.325&0.525
        & 49&\underline{0.21}M&16.87&0.665&0.380
        & \underline{99}&0.84M&14.27&0.517&0.555\\

        \rowrule 4D-Rotor & SIGGRAPH 2024
        & 7&0.35M&16.43&0.352&\underline{0.482}
        & 5&0.38M&\underline{19.48}&\underline{0.738}&0.315
        & 10&0.71M&14.46&0.533&\underline{0.490}\\

        \rowrule RealTime4DGS & ICLR 2024
        & 11&0.38M&16.02&0.359&0.545
        & 12&0.46M&17.53&0.652&0.434
        & 21&1.09M&14.33&0.521&0.553\\

        \rowrule Grid4D & NeurIPS 2024
        & 16&1.15M&16.35&0.334&0.538
        & \underline{80}&\underline{0.21}M&18.30&0.709&0.391
        & 16&\underline{0.46}M&12.47&0.492&0.566\\

        \rowrule Ex4DGS & NeurIPS 2024
        & \underline{61}&2.98M&\underline{16.74}&\underline{0.373}&0.523
        & 32&0.57M&19.21&0.730 &\underline{0.304}
        & 49&1.36M&15.36&0.565&0.531\\

        \rowrule 4C4D & CVPR 2026
        & 52&4.06M&16.69&0.352&0.520
        & 25&4.20M&16.62&0.653&0.424
        & 35&3.02M&\underline{16.99}&\underline{0.614}&0.507\\

        \rowcolor{pink!40}
        \rowrule 
        \projecttitle~(Ours) & -
        & \textbf{127}&\textbf{0.19}M&\textbf{16.84}&\textbf{0.480}&\textbf{0.420}
        & \textbf{246}&\textbf{0.05}M&\textbf{20.28}&\textbf{0.748}&\textbf{0.296}
        & \textbf{214}&\textbf{0.16}M&\textbf{17.42}&\textbf{0.633}&\textbf{0.442}\\

        \midrule

        \rowrule SparseCam4D & CVPR 2026
        & 11&0.30M&23.92 &0.733 &0.251 
        & 8&0.39M&21.33 &0.747 &0.289
        & 12&1.02M&19.21 &0.674 &0.355\\

        \rowcolor{pink!40}
        \rowrule  
        \projecttitle~(Ours) * & -
        & \textbf{123}&\textbf{0.12}M &\textbf{24.87}&\textbf{0.744}&\textbf{0.229}
        & \textbf{263}&\textbf{0.04}M &\textbf{22.04}&\textbf{0.789}&\textbf{0.258}
        & \textbf{228}&\textbf{0.13}M &\textbf{20.25}&\textbf{0.692}&\textbf{0.315}
        \\

        \bottomrule[2pt]
    \end{tabular}
}
\label{tab:main_exp}
\end{table*}

\begin{figure*}[ht!]
    \centering
    \includegraphics[width=1.0\textwidth, trim=0cm 10.5cm 6.5cm 0cm, clip]{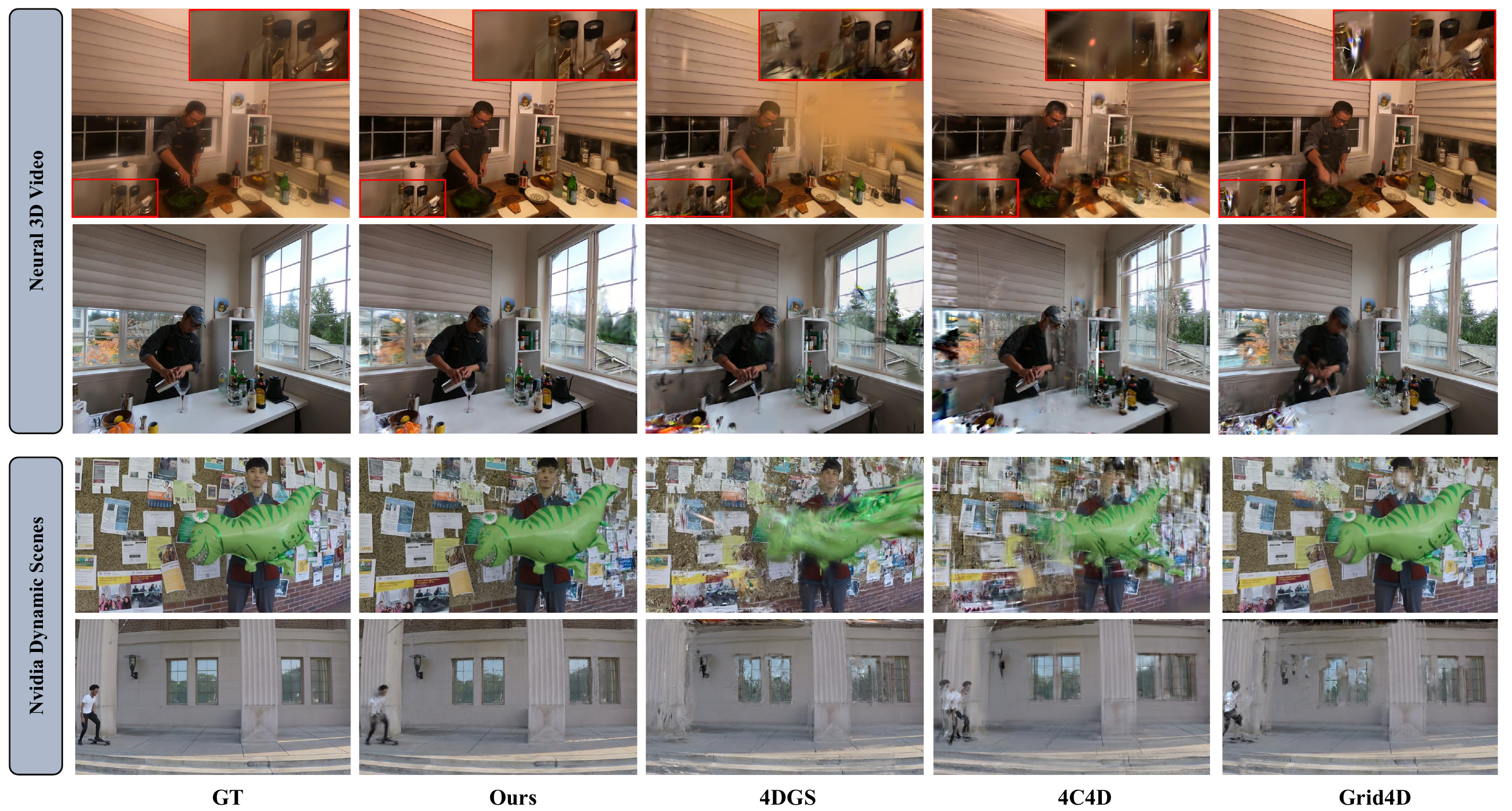}
    \vspace{-0.5cm}
    \caption{\textbf{Qualitative comparisons of state-of-the-art methods.} Visual differences are highlighted with red insets for better clarity. Our approach outperforms other models on different scenes, demonstrating advantages in challenging scenarios. Best viewed in color.}
    \label{fig: comparison}
    \vspace{-0.3cm}
\end{figure*}

\section{Experiments}
\label{Experiments}

\subsection{Experiment Setup}

\paragraph{Datasets.}
Following previous works~\cite{pan2026sparsecam4d, li2024spacetime}, we evaluate the proposed \projecttitle~ on three widely-used sparse-view 4D reconstruction benchmarks: Neu3d \cite{li2022neural}, Nvidia Dynamic \cite{yoon2020novel} and Google Immersive Dataset~\cite{broxton2020immersive} dataset.

\paragraph{Evaluation Metrics.} 
Following standard practices for novel view synthesis, we employ established quantitative metrics to assess rendering quality, including \textbf{PSNR}, \textbf{SSIM}~\cite{wang2004image}, and \textbf{LPIPS}~\cite{zhang2018unreasonable}. 
We additionally report the temporal-PSNR (\textbf{tPSNR}) to evaluate the temporal consistency for the ablation study. 
To further evaluate rendering efficiency, we report the number of Gaussians ($\mathbf{N_{GS}}$) and rendering frames per second (\textbf{FPS}).

\paragraph{Baselines.}
We compare DUET4D with representative dynamic Gaussian reconstruction methods, including 4DGS~\cite{wu20244d}, 
4D-Rotor~\cite{duan20244d}, 
RealTime4DGS~\cite{yang2024real}, 
Ex4DGS~\cite{lee2024fully}, 
and Grid4D~\cite{xu2024grid4d}, as well as recent sparse-view methods 
4C4D~\cite{zhou20264c4d}, 
and SparseCam4D~\cite{pan2026sparsecam4d}. 
For methods whose official sparse-view results are unavailable, we reproduce them using the same input cameras and evaluation protocol. Notably, SparseCam4D adopts a dedicated preprocessing pipeline. To exclude improvements caused by data processing, we additionally construct a strictly controlled comparison.  
Specifically, DUET4D$^{*}$ uses exactly the same preprocessed inputs as SparseCam4D, while all model components and optimization settings remain unchanged.

\subsection{Evaluation}

\paragraph{Quantitative Comparison.}
As reported in Table~\ref{tab:main_exp}, DUET4D consistently achieves the highest average PSNR across all evaluated datasets. 
This performance is particularly notable in dynamic scenes exhibiting complex, non-rigid deformations. 
The superiority of our approach is attributed to the self-corrective coupled guidance, which effectively constrains the restoration process and filters out unreliable pseudo-supervision. Consequently, DUET4D significantly boosts PSNR while maintaining highly competitive SSIM and LPIPS scores. 
Furthermore, to ensure a fair evaluation against SparseCam4D, we adopt an identical preprocessing pipeline. 
Our method not only outperforms SparseCam4D in reconstruction fidelity but also operates at a significantly higher frame rate. 
\paragraph{Qualitative Comparison.} As illustrated in Fig.~\ref{fig: comparison}, whereas baseline methods inherently suffer from incomplete geometry, blurred textures, or floating artifacts in heavily occluded dynamic regions, \projecttitle~produces sharper and temporally coherent renderings. 
This robustness validates that our core designs—sequential anchor alignment, latent steering, and fidelity resampling—collectively cooperation to suppress frame-wise drift, preserve high-frequency details, and eliminate weakly supported primitives.

\paragraph{Efficiency and Compactness.}
Beyond synthesis quality, DUET4D demonstrates a significant advantage in both rendering efficiency and Gaussian compactness. 
As detailed in Table~\ref{tab:main_exp}, our method achieves up to a $49\times$ acceleration in rendering speed compared to state-of-the-art baselines, remarkably utilizing only about $20\%$ of their Gaussian primitive count on Neural 3D dataset. 
The compactness is driven by the proposed Sequential Fidelity Resampling strategy. 
Rather than blindly proliferating primitives to fit dynamic regions, this mechanism allocates capacity: it densifies areas that are repeatedly under-reconstructed while aggressively pruning primitives that are persistently inconsistent or redundant across time. 
This dynamic regularization balanced the trade-off between quality and speed, reducing both the storage overhead and rasterization cost without compromising the fidelity of the reconstructed 4D dynamics.

\subsection{Ablation Study}

Table~\ref{tab:ablation} evaluates the contribution of each proposed component on the Neural 3D Video dataset. To disentangle the effectiveness of the two complementary aspects of DUET4D, we organize the ablation study into representation-side optimization and diffusion-guidance variants. 
We first disable Sequential Anchor Alignment, the framework degrades into one-way from diffusion to Gaussian.
We then employ vanilla single-view diffusion as an intermediate baseline, followed by Anchor-Driven Latent Steering. 
As for the representation side, we disable Sequential Fidelity Resampling and Smoothness Control Regularization to evaluate their effects on reconstruction quality, representation compactness, and rendering efficiency.

\begin{table}[t]
\centering
\caption{\textbf{Ablation study of proposed components on the Neural 3D Video dataset}.
We progressively evaluate Sequential Fidelity Resampling, Smoothness Control Regularization, Sequential Anchor Alignment, and Anchor-Driven Latent Steering.
Vanilla single-view diffusion is included as an intermediate baseline to distinguish the effect of the generative prior from that of the proposed guidance mechanisms.}
\label{tab:ablation}
\resizebox{\columnwidth}{!}{
\setlength{\tabcolsep}{2pt} 
\begin{tabular}{l|cccccc}
\toprule
\textbf{Configuration} 
& \textbf{FPS} 
& $\textbf{N}_{\text{GS}}$ 
& \textbf{tPSNR$\uparrow$}
& \textbf{PSNR$\uparrow$}
& \textbf{SSIM$\uparrow$}
& \textbf{LPIPS$\downarrow$} \\
\midrule
Baseline
& 80 & 0.21M & 36.91 & 18.30 & 0.709 & 0.391 \\

\midrule

+Fidelity Resampling
& 268 & 0.04M & 36.89 & 18.54 & 0.713 & 0.358 \\

+Control Regularization
& 271 & 0.03M & 36.88 & 18.80 & 0.729 & 0.329 \\

\midrule

+Vanilla Diffusion
& 230 & 0.06M & 36.33 & 20.04 & 0.742 & 0.297 \\

+Anchor Alignment 
& 233 & 0.06M & 36.89 & 20.08 & 0.743 & 0.305 \\

+Latent Steering (Full)
& 246
& 0.05M
& 37.99
& 20.28
& 0.748
& 0.296 \\

\bottomrule
\end{tabular}
}
\end{table}

\begin{figure}[t]
    \centering
    \includegraphics[width=1\linewidth, trim=1cm 5.9cm 2.3cm 0cm, clip]{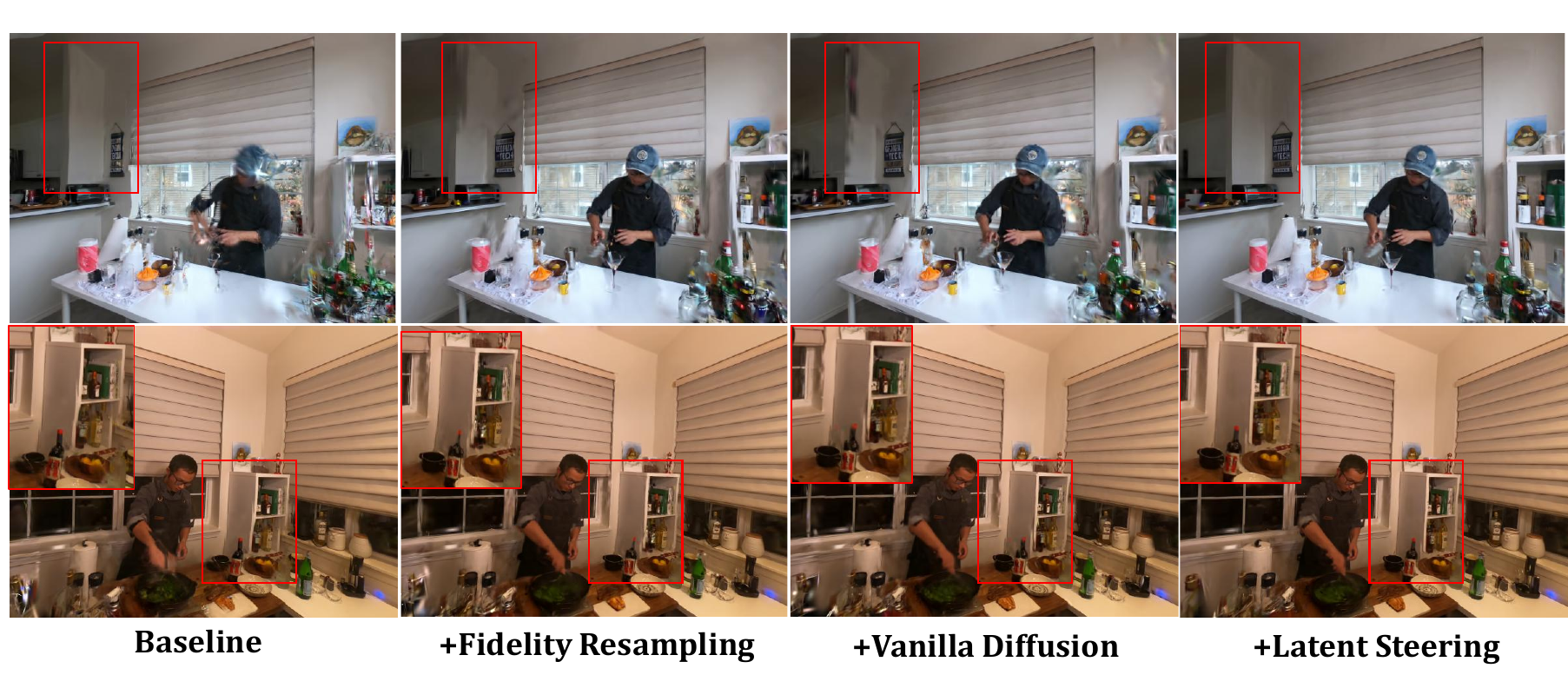}
    \caption{\textbf{Visual ablation.} Fidelity Resampling removes floating artifacts, while vanilla diffusion restores details but may introduce appearance drift. The full model uses Gaussian-derived anchors and confidence-aware latent steering to recover sharper and more consistent results.}
    \label{fig:visual_ablation}
\end{figure}

\paragraph{Effectiveness of Anchor Alignment.}
Compared with vanilla single-view diffusion, Sequential Anchor Alignment improves temporal PSNR and rendering quality. 
This clear temporal improvement verifies that Gaussian-derived correspondences provide effective constraints across neighboring views and frames, allowing independently restored results to share consistent appearance corrections. By aligning restoration residuals rather than directly warping complete RGB predictions, the module reduces frame-wise restoration drift without overriding the geometry and view-dependent content encoded by the current representation. 
The aligned residuals therefore establish a more reliable temporal condition for the subsequent latent guidance stage.

\paragraph{Effectiveness of Fidelity Resampling.}
Sequential Fidelity Resampling substantially increases representation compactness and rendering efficiency. 
The results indicate that confidence-gated multi-view evidence can distinguish repeatedly supported reconstruction deficits from primitives that remain inconsistent with the available observations. 
As a result, the method allocates new capacity only to reliably under-reconstructed regions while removing weakly supported or redundant Gaussians.

\paragraph{Effectiveness of Smoothness Control.}
Adding Smoothness Control Regularization further increases rendering quality. The Gaussian count is further reduced from 0.04M to 0.03M, and the rendering speed remains nearly unchanged. This result suggests that the spatial and temporal smoothness constraints prevent the deformation field from fitting local noise. They improve reconstruction quality while preserving a compact and efficient representation.

\begin{figure}
    \centering
    \includegraphics[width=1\linewidth, trim=5cm 1.1cm 5.5cm 0cm, clip]{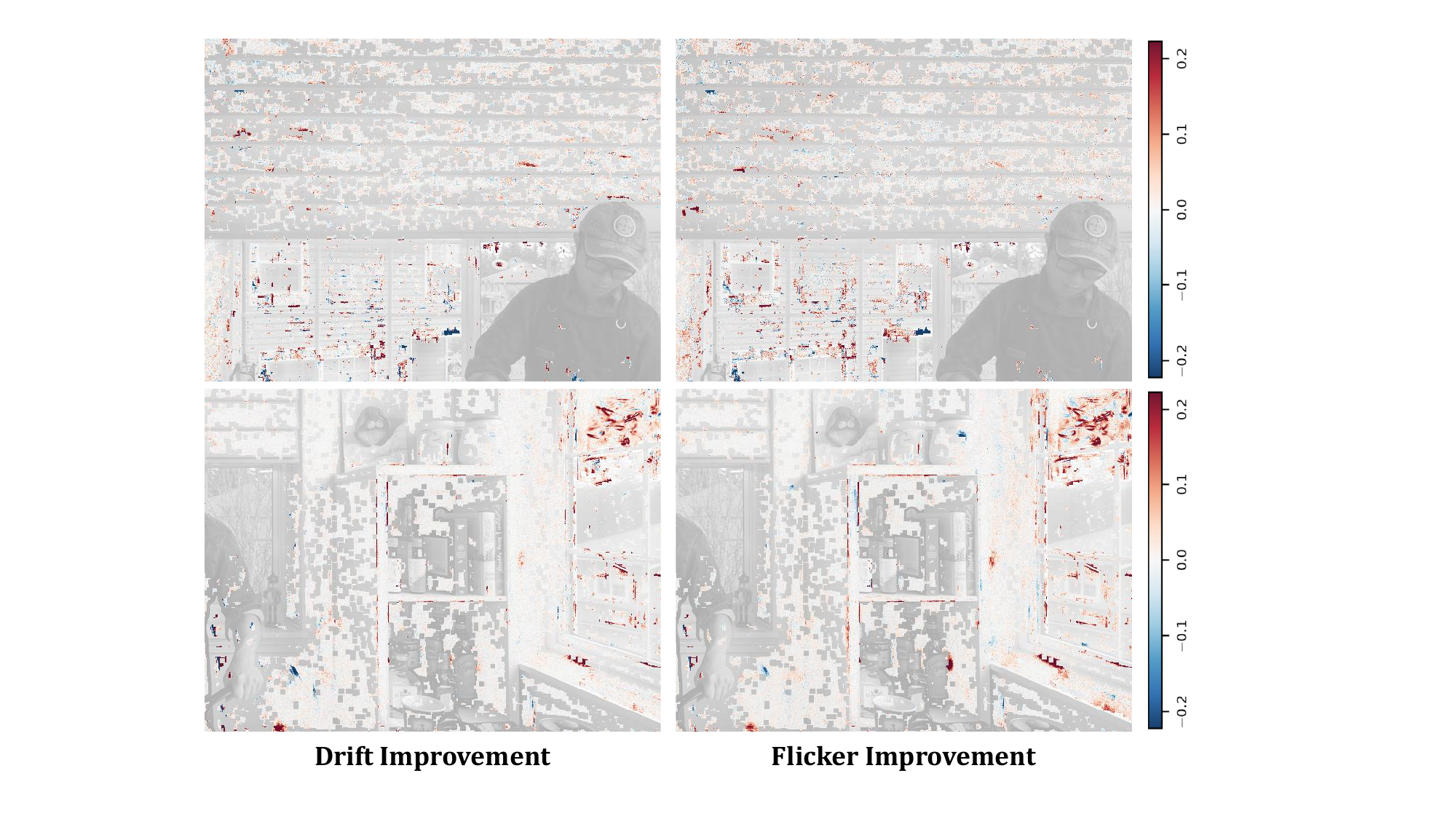}
    \caption{\textbf{Temporal consistency on static regions.} We compare vanilla single-view diffusion with the full method on fixed-view sequences. The left column shows the difference in long-term drift from a reference frame, while the right column shows the difference in short-term flicker between adjacent frames. Red indicates lower error for \projecttitle, and blue indicates the opposite.}
    \label{fig:visual_ablation_temporal}
    \vspace{-0.3cm}
\end{figure}

\paragraph{Effectiveness of Latent Steering.}
Anchor-Driven Latent Steering yields the best overall performance. 
The gains validate the use of aligned evidence to steer the diffusion prediction in latent space rather than directly blending restored RGB images. The guidance improves temporal coherence while preserving local details, leading to more accurate and stable pseudo-supervision.

\paragraph{Visual Ablation.}
Figure~\ref{fig:visual_ablation} compares the main variants in weakly observed regions. Fidelity Resampling removes weakly supported Gaussians and improves geometry, but cannot recover missing appearance. Vanilla single-view diffusion restores sharper textures, but its independent predictions cause color drift and incorrect details. Our full model uses Gaussian-derived anchors to align restoration across views and time. Confidence-aware latent steering further limits corrections to regions with reliable correspondence. 

Figure~\ref{fig:visual_ablation_temporal} evaluates temporal stability on fixed-view sequences. We measure long-term drift from a reference frame and short-term flicker between adjacent frames in static regions. The full model reduces both errors, with clear improvements near the cabinet and window. These results confirm that Gaussian-derived correspondences provide effective temporal constraints, while confidence-aware steering suppresses unsupported corrections. Together, these components improve both spatial fidelity and temporal consistency, providing cleaner and more stable pseudo-supervision for 4D reconstruction.

\paragraph{Sensitive Analysis.} Table~\ref{tab:hyperparameter_sensitivity} evaluates sensitivity of \projecttitle~to the guidance strength and the confidence threshold. Moderate parameter choices provide a better balance between temporal consistency and spatial fidelity. Excessive latent guidance may over-constrain the restoration process and weaken local appearance details. Similarly, a loose confidence threshold may introduce unreliable pseudo-supervision, while a strict threshold may discard useful correspondence evidence. The performance remains stable within a reasonable range, demonstrating the robustness of \projecttitle~and supporting the system design.

\begin{table}[t]
\centering
\caption{\textbf{Hyperparameter sensitivity on the Neural 3D Video dataset.} For each group, only the indicated parameter is changed while all other settings remain fixed. }
\label{tab:hyperparameter_sensitivity}
\resizebox{0.46\textwidth}{!}{
\begin{tabular}{l|l|ccccc}
\toprule
  & \textbf{Setting}
& \textbf{PSNR}$\uparrow$
& \textbf{SSIM}$\uparrow$
& \textbf{LPIPS}$\downarrow$
& \textbf{tPSNR}$\uparrow$
 \\

\midrule
\multirow{3}{*}{$\rho$} 
& $0.25$ & 20.21 & 0.744 & 0.299 & 37.82  \\
& $0.50$ & \textbf{20.28} & \textbf{0.748} & \textbf{0.296} & 37.99  \\
& $1.00$ & 20.06 & 0.739 & 0.325 & \textbf{38.32}  \\
\midrule
\multirow{4}{*}{$\tau_m$} 
& $0.001$ & 20.18 & 0.742 & 0.300 & 37.79  \\
& $0.002$ & \textbf{20.28} & \textbf{0.748} & \textbf{0.296} & \textbf{37.99}  \\
& $0.005$ & 20.22 & 0.744 & 0.298 & 37.87  \\
& $0.010$ & 20.12 & 0.737 & 0.302 & 37.61  \\
\bottomrule
\end{tabular}
}
\end{table}

\subsection{Limitations}

DUET4D relies on the current Gaussian representation to construct spatiotemporal correspondences. When the initial geometry is severely corrupted, or when an object undergoes extremely fast motion and long-term occlusion, the available Gaussian evidence may be insufficient to produce reliable anchors. 
In such regions, the confidence mechanism conservatively falls back to independent single-view restoration, which limits error propagation but cannot fully recover content that is never observed.
Future work may explore long-range correspondence aggregation and more efficient on-demand pseudo-view scheduling.

\section{Conclusion}
We presented \projecttitle, a self-corrective framework for sparse-view 4D Gaussian reconstruction. 
This design is motivated by how to mitigate the spatiotemporal inconsistency inherent in the generative prior to improve the reconstruction quality.
By using the evolving Gaussian representation to guide temporally coherent restoration with a frozen single-view diffusion model, and selectively converting confidence-filtered discrepancies into Gaussian densification and pruning policies, 
\projecttitle~ enables the generative prior and explicit representation to progressively refine each other. 
Extensive experiments on three widely-used datasets demonstrate that our method improves novel-view rendering quality while maintaining high rendering efficiency and a compact Gaussian representation.

\bibliography{aaai2027}

\end{document}